%% file: main.tex
\documentclass{article}

\usepackage{microtype}
\usepackage{graphicx}
\usepackage{subfigure}
\usepackage{booktabs} %

\usepackage{hyperref}

\usepackage[accepted]{icml2024}

\usepackage{amsmath}
\usepackage{amssymb}
\usepackage{mathtools}
\usepackage{amsthm}

\usepackage[capitalize,noabbrev]{cleveref}

\theoremstyle{plain}

\theoremstyle{definition}

\theoremstyle{remark}

\input{math-notation}

\usepackage[textsize=tiny]{todonotes}
\usepackage{verbatim}

\newcommand{\codeURL}{\url{https://github.com/AustinT/basic-mol-bo-workshop2024}}
\icmltitlerunning{
Diagnosing and fixing common problems in Bayesian optimization for molecule design
}

\begin{document}

\twocolumn[
\icmltitle{%
Diagnosing and fixing common problems in Bayesian optimization \\ for molecule design
}

\icmlsetsymbol{equal}{*}

\begin{icmlauthorlist}
\icmlauthor{Austin Tripp}{camb}%
\icmlauthor{Jos\'e Miguel Hern\'andez-Lobato}{camb}
\end{icmlauthorlist}

\icmlaffiliation{camb}{Department of Engineering, University of Cambridge, Cambridge, UK}

\icmlcorrespondingauthor{Austin Tripp}{ajt212[at]cam.ac.uk}

\icmlkeywords{Machine Learning, ICML}

\vskip 0.3in
]

\printAffiliationsAndNotice{\icmlEqualContribution} %

\begin{abstract}
Bayesian optimization (BO) is a principled approach to molecular design tasks.
In this paper we explain three pitfalls of BO which can cause poor empirical performance:
an incorrect prior width, over-smoothing, and inadequate acquisition function maximization.
We show that with these issues addressed,
even a basic BO setup
is able to achieve the highest overall
performance on the PMO benchmark for molecule design
\citep{gao2022sample}.
These results suggest that BO may benefit from more attention
in the machine learning for molecules community.
\end{abstract}

\section{Introduction}

Many problems in drug discovery can be summarized
as finding molecules with desirable properties.
This is often formalized as maximizing a property function
$f:\mathcal M\mapsto \mathbb{R}$,
where $\mathcal M$ denotes the space of molecules.
The challenge of this problem is the immense size of the search space $\mathcal M$:
out of an estimated $10^{60}$
possible molecules \citep{bohacek1996art},
only a minuscule fraction can be tested experimentally
(perhaps $10^2$--$10^4$).
Therefore, algorithms for molecule design must operate very efficiently,
making the best use of their experimental budget.

Despite the need for efficiency,
the current most popular algorithms for molecule design all seem to heavily rely on
\emph{random} exploration.
Genetic algorithms (GAs) and their variants randomly mutate and combine known molecules
\citep{jensen2019graph,Nigam2020Augmenting}.
Algorithms based on reinforcement learning (RL) such as REINVENT
\citep{olivecrona2017molecular,blaschke2020reinvent}
and GFlowNets \citep{bengio2021flow,bengio2023gflownet}
instead make random perturbations to a molecule generation policy.
In both cases, because the exploration is random
it is likely to be inefficient.

In contrast,
Bayesian optimization (BO)
stands out as a principled alternative which performs \emph{deliberate}
exploration \citep{garnett_bayesoptbook_2023}.
By explicitly using prior knowledge to model molecular properties,
BO algorithms can make a precise trade-off between
exploration (testing new molecules)
and exploitation (testing molecules similar to the best known ones).
Because of this, one might expect BO methods to be state-of-the-art
in this field.
Surprisingly however, prior work has shown that BO under-performs RL/GA methods \citep{gao2022sample}.

In this short paper, we argue that poor BO performance in prior works may
essentially be due to poor tuning of hyperparameters.
To show this, we first introduce BO (\S\ref{sec:background})
and explain several ways in which certain choices of hyperparameters
can lead to \emph{predictably} poor optimization performance (\S\ref{sec:pitfalls}).
Second, we show that with the right settings
a basic BO setup achieves the best reported performance
on the PMO benchmark for molecular optimization algorithms \citep{gao2022sample}.
We conclude with a brief evaluation of the pros and cons of BO,
arguing that while it is not perfect, it should likely
receive more attention from the community (\S\ref{sec:discussion}).

\section{Background on Bayesian optimization}
\label{sec:background}

Let $\X$ represent an input space.
Let $\prob$ denote the probability of an event,
$\E$ denote expected value,
and $\V$ denote variance.
The most basic form of
Bayesian optimization (BO) seeks
\begin{equation}
    x^*=\argmax_{x\in\X} f(x)\ ,
\end{equation}
namely an input which maximizes
an \emph{objective function}
$f:\X\mapsto\R$.
At the heart of BO is a \emph{probabilistic surrogate model},
which specifies a \emph{distribution} over surrogate models
$\hat f: \X\mapsto\R$ for the objective function $f$.
We will denote a general probabilistic surrogate model
by $p(\hat f)$.

BO uses $p(\hat f)$ to choose inputs to evaluate,
typically choosing an input $x$ which maximizes
an \emph{acquisition function} $\alpha$.
An intuitive example of an acquisition function is the \emph{probability of improvement} (PI) \citep[\S7.5]{garnett_bayesoptbook_2023}
\begin{equation}\label{eq:pi-acq-fn}
    \alpha_{\text{PI}}\left(x;p(\hat f),y_{\text{best}}\right)
    = \prob_{\hat f\sim p(\hat f)}\left[ \hat f(x) > y_{\text{best}} \right]\ ,
\end{equation}
which measures the probability that $f(x)$ will improve upon the incumbent best
measurement $y_{\text{best}}$:
an intuitively reasonable criterion to select points for evaluation.

Pseudocode for a general BO loop is given in 
Algorithm~\ref{alg:general-bayesopt}.
The key lines of this algorithm are line~\ref{alg:bo:fit model}
(which defines the probabilistic surrogate model)
and line~\ref{alg:bo:maximise acqn fn}
(which uses an acquisition function to select an input to evaluate).\footnote{
    To allow the acquisition function to vary over iterations,
    we use the notation $\alpha_i$.
}
The rest of this section will discuss these steps in more detail.

\begin{algorithm}[h]
\caption{General Bayesian optimization loop.}
\label{alg:general-bayesopt}
\begin{algorithmic}[1]
\REQUIRE Input dataset $\D_0=\{(x_1,y_1),\ldots,(x_n,y_n)\}$, acquisition function $\alpha$
\FOR{$i$ in $1,2,\ldots$}
    \STATE\label{alg:bo:fit model} Fit $p_i(\hat f)$ to dataset $\D_{i-1}$
    \STATE\label{alg:bo:maximise acqn fn} Select $x_i=\argmax_{x} \alpha_i(x;p_i(\hat f))$\hfill%
    \STATE Acquire label $y_i$ for $x_i$
    \STATE $\D_i\gets \D_{i-1}\cup\{(x_i,y_i)\}$\hfill%
    \IF{computational budget is exhausted}
        \RETURN $\D_i$\hfill\COMMENT{Terminate}
    \ENDIF
\ENDFOR
\end{algorithmic}
\end{algorithm}

\subsection{Gaussian process surrogate models}

Gaussian processes (GPs) are the most commonly-used
class of probabilistic surrogate models,
and therefore we will introduce them briefly here.
A GP assumes the that \emph{joint} distribution
of the observed data is Gaussian,
whose mean is given by a mean function
$\mu:\X\mapsto\R$,
and whose covariance is given by a positive-definite
\emph{kernel function}
$k:\X\times\X\mapsto\R$.
When $\X=\R^n$,
a common choice of kernel function
is the RBF kernel,
defined as
\begin{equation}\label{eqn:rbf-kernel}
    k_{\text{RBF}}(x,x')=\sigma^2 \exp{\left(\frac{-\|x-x'\|^2}{2 \ell^2}\right)}\ .
\end{equation}
The hyperparameter $\sigma$ is referred to as the \emph{kernel amplitude}
(because the marginal prior distribution for every input
is a Gaussian with standard deviation $\sigma$),
while $\ell$ is referred to as the \emph{lengthscale}.

The primary appeal of GP models is that their posterior distribution
has an analytic solution,
evading the need for approximate inference techniques like MCMC.
The formulas for the analytic solution
can be found in numerous textbooks
\citep{rasmussen2006gp,garnett_bayesoptbook_2023}.
This allows the model fitting step
in line~\ref{alg:bo:fit model} to be performed efficiently
and reliably.

GP surrogate models will be used in the remainder of this paper.
However, we emphasize that BO does not \emph{require}
the use of GP models:
Bayesian neural networks or ensembles are viable alternatives.

\subsection{Acquisition functions}

Despite its simplicity,
the PI acquisition function in equation~\ref{eq:pi-acq-fn}
is seldom used in practice,
chiefly because it 
does not account for the \emph{magnitude} of the improvement
(so large improvements are treated the same as small improvements).
Instead, many people use
\emph{expected improvement} (EI)
\begin{equation}
    \alpha_{\text{EI}}\left(x;p(\hat f),y_{\text{best}}\right)
    = \E_{\hat f\sim p(\hat f)}\left[ \max\left(0, \hat f(x) - y_{\text{best}}\right) \right]
    \ ,
\end{equation}
which measures the average \emph{amount} by which $f(x)$ is predicted to improve over $y_{\text{best}}$.
Another common acquisition is the \emph{upper confidence bound} (UCB)
\begin{equation}
    \alpha_{\text{UCB}}\left(x;p(\hat f)\right)
    = \E_{\hat f}\left[ \hat f(x) \right]
    + \beta \sqrt{\V_{\hat f}\left[\hat f(x)\right]}
    \ ,
\end{equation}
which is the mean prediction plus the standard deviation weighted by $\beta$.
There are many other choices of acquisition function:
\citet[Chapter 7]{garnett_bayesoptbook_2023} gives a good introduction to them.

Importantly, the acquisition function is \emph{not} something which should be chosen arbitrarily.
Because the acquisition function specifies (implicitly or explicitly) the explore-exploit trade-off,
it should be chosen with that in mind.
In general, EI tends to be \emph{exploitative},
while UCB becomes more exploitative as $\beta\to0$ and more exploratory
as $\beta\to\infty$.

\section{Common Bayesian optimization pitfalls}
\label{sec:pitfalls}

\begin{figure}[h]
    \centering
    \includegraphics{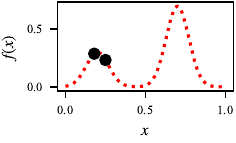}
    \caption{
        1D optimization task meant to be qualitatively similar to molecular design tasks.
        Only a small number of data points are known (black dots), none of which are near the global optimum of the unknown function (red dashed line).
    }
    \label{fig:1D-bo}
\end{figure}

While there is are no universal rules to optimally tune all hyperparameters in BO,
some hyperparameter settings have intuitive and predictable failure modes.
We explain three possible failure modes
with an illustrative example in 1D,
shown in Figure~\ref{fig:1D-bo}.
This setup is chosen to be vaguely analogous to molecule design:
some molecules near a local optimum are known,
but other more promising optima are unexplored.
We use a GP with an RBF kernel as the surrogate model
(typically the default choice in most GP libraries)
with low observation noise.

\subsection{Pitfall \#1: prior width}
\label{pitfall:prior width}

\begin{figure}[tb]
    \centering
    \includegraphics{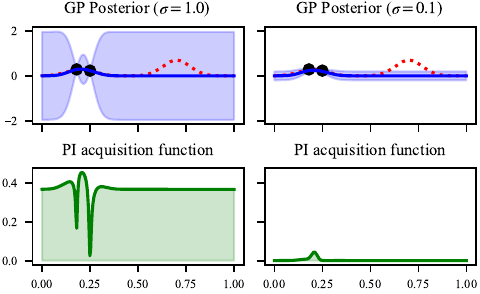}
    \caption{
        Effect of prior width parameter $\sigma$ in a GP model,
        illustrating ``prior width'' pitfall (\S\ref{pitfall:prior width}).
        Low values of $\sigma$ cause the model to predict lower returns from exploration.
    }
    \label{fig:prior-width}
\end{figure}

A model $p(\hat f)$ will imply a range of values that $f$
is likely to take, which we refer to as the \emph{prior width}.
For example, with a GP model, the predictive standard deviation
can be interpreted as a prior width,
and can be controlled by the parameter $\sigma$ (equation~\ref{eqn:rbf-kernel}).
The prior width directly determines the predicted gains from exploring
away from the training data.
Figure~\ref{fig:prior-width} directly shows the consequence of this,
using prior widths of $0.1$ and $1.0$.
When $\sigma=0.1$,
the points near the left are predicted to be nearly optimal,
and there is no predicted gain from exploring the right side of the space.
In contrast, when $\sigma=1.0$, the points near the right have a reasonably high predicted probability
of being better than the points on the left.

It is straightforward to see that the same principles will also hold
outside of 1D examples.
A general guideline is that if $\sigma$ is too high, then BO algorithms
will anticipate large gains from exploration
and tend to be too exploratory.
Conversely, if $\sigma$ is too low then BO algorithms will under-explore.

\subsection{Pitfall \#2: over-smoothing}
\label{pitfall:over smoothing}

\begin{figure}[tb]
    \centering
    \includegraphics{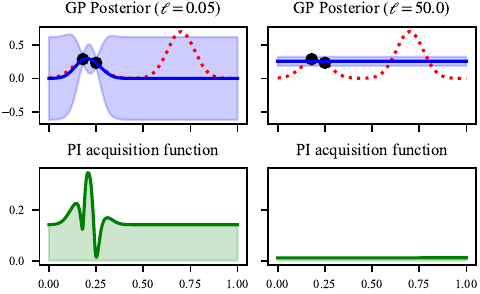}
    \caption{
        Effect of lengthscale parameter $\ell$ in a GP model,
        illustrating ``over-smoothing'' pitfall (\S\ref{pitfall:over smoothing}).
        High values of $\ell$ also imply lower returns from exploring
        inputs near known inputs.
    }
    \label{fig:over-smoothing}
\end{figure}

The probabilistic surrogate model $p(\hat f)$
essentially encodes how measurements of known input points
influence those of unmeasured points.
For GPs in 1D, each point can be thought of as
having a ``radius'' of influence around it,
which is determined by the lengthscale of the kernel function
(e.g.\@ $\ell$ in equation~\ref{eqn:rbf-kernel}).
If this radius is too high,
it can lead to overconfident predictions.
Figure~\ref{fig:over-smoothing} illustrates this by showing
the GP posterior using an RBF lengthscale of $\ell=0.05$
and $\ell=5.0$.
When $\ell=50.0$,
the measurements on the left 
suggest that the right side is not worth exploring,
which does not happen when $\ell=0.05$.

A general guideline is that over-smoothing will result in under-exploration,
while under-smoothing will result in over-exploration.

\subsection{Pitfall \#3: inadequate search}

Line~\ref{alg:bo:maximise acqn fn} requires finding an input
which maximizes the acquisition function.
Although in 1D this can be accomplished via a comprehensive grid search,
in combinatorially large spaces like molecules
inevitably only a small fraction of all candidate points may be considered.
Unfortunately, popular search methods like generative models and GAs
tend to propose molecules similar to known molecules.
In 1D, this is a bit like only searching in a narrow interval around the known points,
akin to never considering inputs on the right side of Figure~\ref{fig:1D-bo}.

Unlike the first two pitfalls, poor search should only ever result in \emph{under}-exploration.
However, longer searches will generally take more time.

\section{Experiments: fixing these issues substantially improves performance}
\label{sec:experiments}

\begin{figure}
    \centering
    \includegraphics[width=\linewidth,trim={0 400px 0 450px},clip]{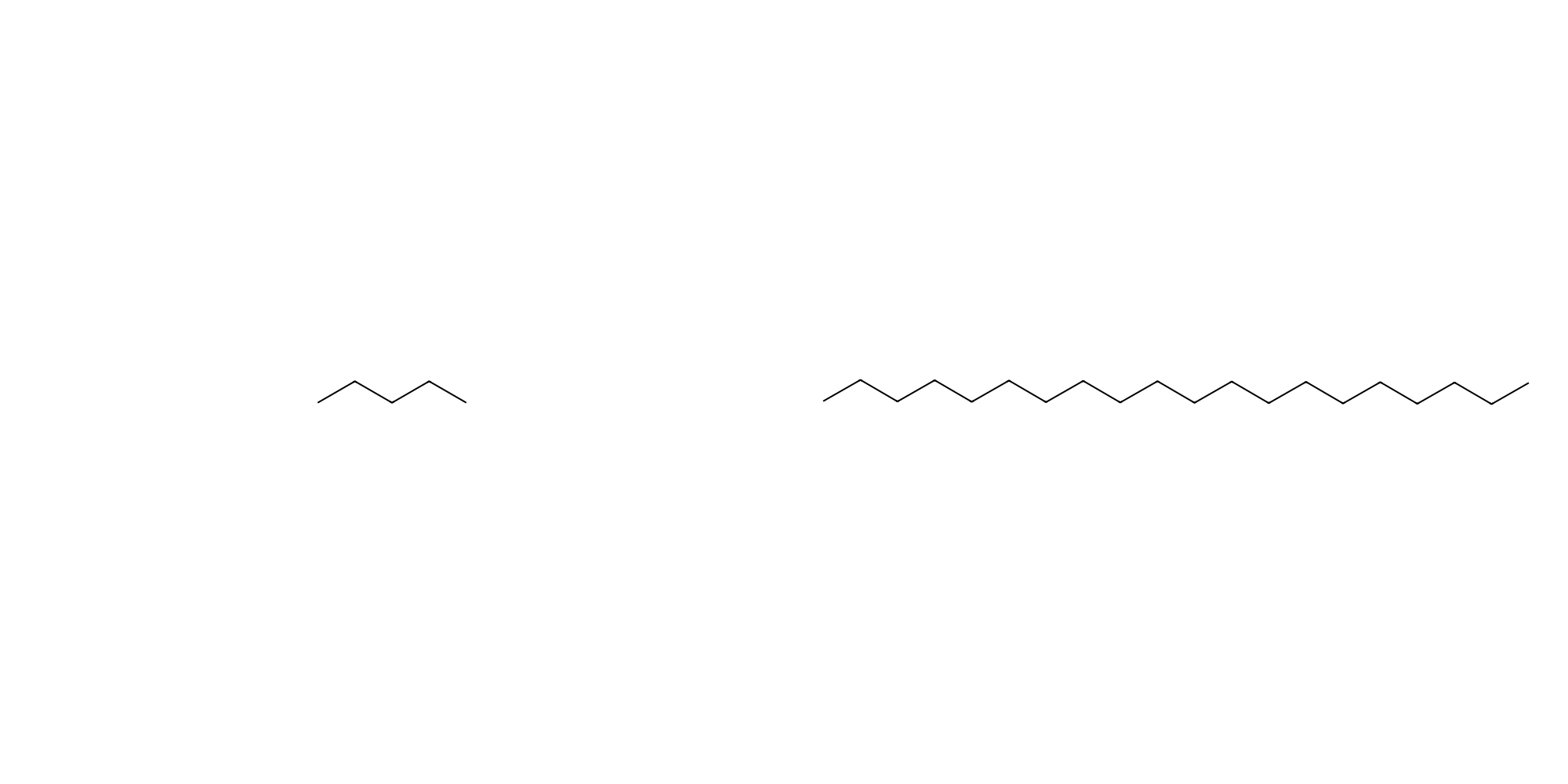}
    \includegraphics[width=\linewidth,trim={0 200px 0 0}]{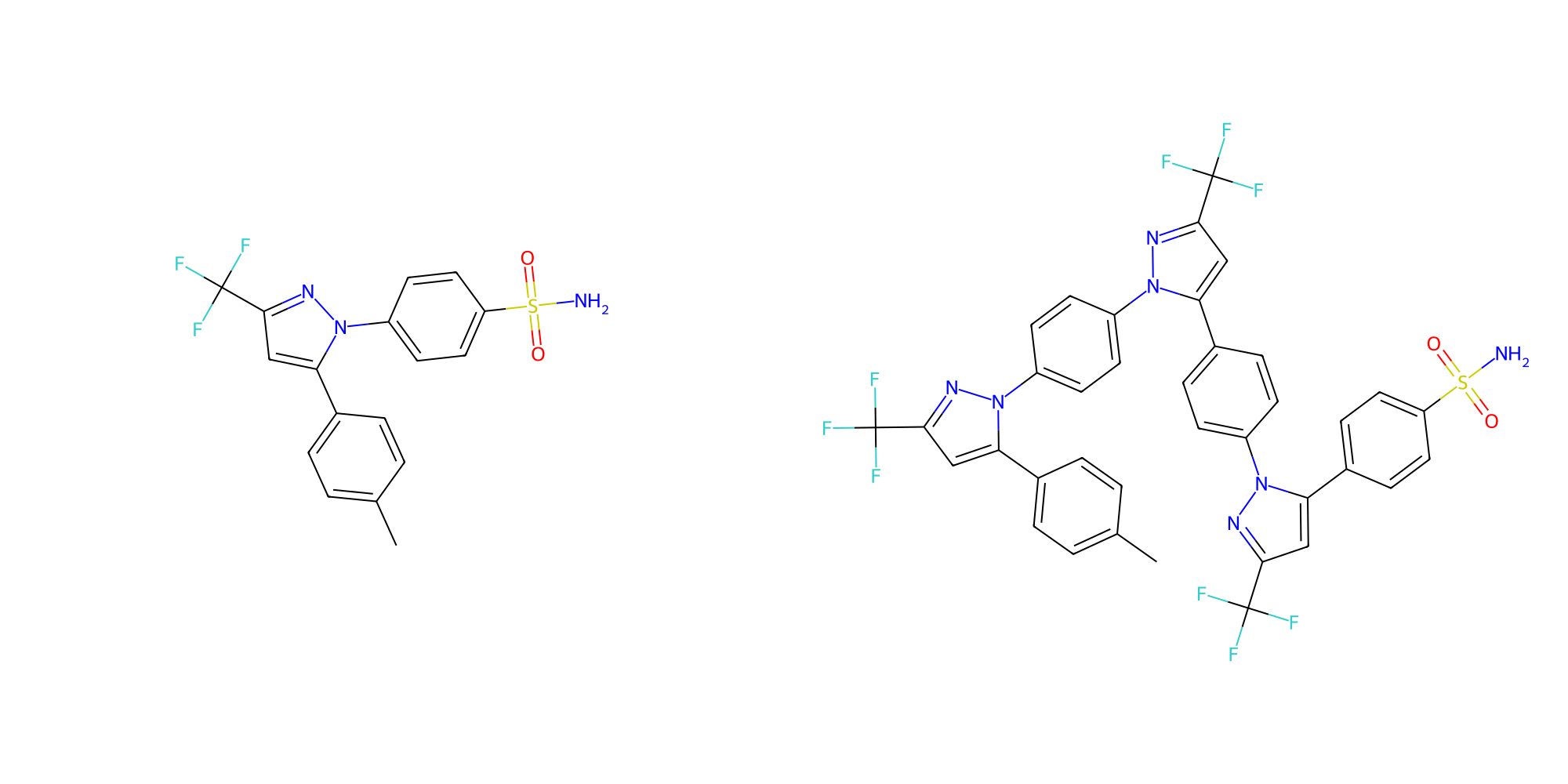}
    \caption{
        Two pairs of molecules whose \emph{binary} Morgan fingerprints of radius 2 are identical.
        The top pair is two alkanes of different lengths,
        which only contain \texttt{-CH$_3$} and \texttt{-CH$_2$-} groups.
        The bottom pair is the anti-inflammatory drug molecule
        Celecoxib
        and a larger analogue with many repeated substructures.
        SMILES strings are given in Appendix~\ref{apdx:smiles from figure}.
    }
    \label{fig:similar-molecules-binaryfp}
\end{figure}

In this section we consider the application of BO
to the PMO benchmark,
which consists of 23 different objective functions
$f:\mathcal M\mapsto [0,1]$ over molecule space
\citep{gao2022sample}.
Very few works have applied BO to this benchmark,\footnote{
    Aside from \citet{gao2022sample},
    we are only aware of \citet{wang2023graph}.
}
so we focus our attention to the ``GP BO'' baseline
implemented by \citet{gao2022sample}.
Their implementation used a basic Tanimoto kernel on molecular fingerprint features
\begin{equation}\label{eqn:tanimoto kernel}
    k(x,x')=\sigma^2 T\left(\textrm{fp}(x), \textrm{fp}(x')\right)\ ,
\end{equation}
where $T$ denotes the Tanimoto coefficient\footnote{Also called Jaccard similarity} function
and $\textrm{fp}$ is a function producing molecular fingerprints.
They used a UCB acquisition function with random value of $\beta$ in each iteration,
which was optimized using
a Graph GA algorithm
\citep{jensen2019graph}.
However, a close inspection of their implementation
reveals potential signs of all 3 pitfalls
from Section~\ref{sec:pitfalls}:
\begin{enumerate}
    \item \textbf{Prior width}:
        the kernel hyperparameters are chosen by maximizing the marginal likelihood on the \emph{starting} data,
        which mainly consists of molecules with poor scores.
        This is likely to select a lower value of $\sigma$.
    \item \textbf{Over-smoothing}:
        a GP with a Tanimoto kernel over \emph{binary} Morgan fingerprints is used.
        Since binary fingerprints track only the presence or absence of certain structures
        rather than their count, it is possible for molecules of vastly different
        sizes to be judged as highly similar.
        Figure~\ref{fig:similar-molecules-binaryfp} shows some examples.
    \item \textbf{Inadequate search}:
        their Graph GA used a very small number of iterations compared
        to the batch size,
        such that for every molecule chosen only $\approx6$
        molecules were proposed by the GA.
        This is a relatively low number, especially as GAs tend to propose
        molecules which are very similar to the starting molecules.
        Ultimately, this likely resulted in significant under-exploration.
\end{enumerate}

To address these issues, we created a modified implementation of GP BO.
To ensure a suitable prior width,
we set $\sigma=1.0$ for the GP kernel (equation~\ref{eqn:tanimoto kernel})
knowing that all objectives in the PMO benchmark
lie in the interval $[0,1]$.
This ensures that the model assigns a reasonable probability to all possible values.
To fix over-smoothing, we used \emph{count} Morgan fingerprints instead of binary fingerprints.
Finally, to improve the search, we
tuned the genetic algorithm parameters
to propose $\approx 1000$ molecules per molecule chosen.
We also decreased the batch size to $1$ to allow for more iterations.
To keep computational costs reasonable, we only ran BO for $1000$ iterations
($10\%$ of the evaluation budget),
and chose the remaining $9000$ molecules in one large batch
by maximizing the GP posterior mean.
More details and a link to our code can be found in Appendix~\ref{apdx:bo details}.

As recommended by \citet{gao2022sample},
we report the AUC Top-10 metric,
which is the normalized area under the curve of the 10th
best molecule over time.
The AUC Top-10 results from out experiments is shown
in Table~\ref{tab:pmo-benchmark-results}.
The sum of AUC Top-10 scores for our GP BO method
$16.303$
which is not only higher than the best method from \citet{gao2022sample}
(REINVENT, with a score of $14.196$),
but also higher than subsequently reported results
from \citet{tripp2023genetic} and \citet{kim2024genetic}.
Importantly, our GP BO implementation improves upon the implementation
from \citet{gao2022sample} by over 3.0 points,
which is about the same as the score difference
between the best and 10th best methods from \citet{gao2022sample}.
This suggests that our changes did have a significant impact.

\section{Discussion}
\label{sec:discussion}

This short paper surveyed several potential failure modes of BO
(\S\ref{sec:pitfalls})
and showed empirically that a basic BO implementation
with these issues resolved is able to achieve
state-of-the-art performance on the PMO benchmark \citep{gao2022sample}.

However, what this paper presents should best be thought of as a very limited
pilot study, rather than a full diagnosis of potential issues in BO.
Importantly, we \emph{do not} claim that BO will work well
if the three pitfalls we present are avoided.
We also did not perform an ablation study,
and therefore our results do not provide insight into how much
each component of BO influences the overall result.
Additionally, we did not experiment with changing the acquisition function,
which in practice should significantly impact BO behavior.
Finally, it is unclear whether results from single-task, noiseless, and unconstrained
optimization will translate to real-world problems
which tend to be multitask, noisy, and highly constrained.

Nevertheless, we think there are good reasons to continue research
into BO algorithms for molecule design.
Aside from empirical performance,
the BO framework allows domain experts to incorporate their knowledge
into the probabilistic surrogate model,
and produces decisions which are interpretable and correctable.\footnote{
    Specifically, the question of why one decision was made over another
    can be reduced to comparing the model's predictions for each decision,
    making them interpretable.
    If the user dislikes a decision, they can correct it
    by either changing the model (to change its predictions)
    or changing the acquisition function (to change how decisions are made from predictions).
}
These are highly desirable properties for practical molecule design problems.
Improving surrogate models and extending BO to more
complex optimization settings are active research
areas which plausibly still have a lot of low-hanging fruit left.
Overall, we hope the reader concludes from this paper that BO is a promising
technique for molecule design,
and finds the explanations and fixes of common BO problems useful.

\bibliography{references}
\bibliographystyle{icml2024}

\newpage
\appendix
\onecolumn

\section{Details of BO setup}
\label{apdx:bo details}

Full code for our experiments is available at:
\begin{center}
\codeURL{}
\end{center}

Our implementation used:
\begin{itemize}
    \item An initial set of \textbf{10} molecules randomly sampled from the ZINC\ 250k dataset.
    \item A BO batch size of \textbf{1} (i.e.\@ one molecule is selected every iteration)
    \item The default GA from the \textsc{MolGA} package was used as the optimizer.
        It used a population size of $10^4$, an offspring size of $200$, and $5$ generations.
    \item To prevent excessively large molecules from being produced,
        molecules were limited to have at most $100$ heavy atoms.
    \item A UCB acquisition function with random $\beta$ values, (logarithmically) evenly distributed in $[10^{-2}, 10^0]$.
\end{itemize}

To reduce computational requirements, we ran the above procedure for $990$ iterations,
then selected the remaining $9000$ allowable molecules randomly.
This means that our results are likely an \emph{underestimate} of BO's potential.

\clearpage
\section{Full results}

See Table~\ref{tab:pmo-benchmark-results}.
The full results (including dis-aggregated AUC values and log files)
are available at:
\begin{center}
    \codeURL
\end{center}

\begin{table}[h]
  \caption{
    AUC top-10 scores on PMO benchmark \citep{gao2022sample}.\quad
    $^*$Taken from \citet{gao2022sample}.\\
    $^{**}$Taken from \citet{tripp2023genetic}.
    $^\dagger$Taken from \citet{kim2024genetic}.
  }
  \label{tab:pmo-benchmark-results}
  \centering
  \begin{tabular}{c|rrrr}

  \toprule
    Method & REINVENT$^*$ & MolGA$^{**}$ & Genetic GFN$^\dagger$  & Our GP BO \\  
    \midrule
    albuterol\_similarity & 0.882 $\pm$ 0.006 &0.896 $\pm$ 0.035&{0.949 {$\pm$ 0.010}}&0.964 $\pm$ 0.050\\
    amlodipine\_mpo & 0.635 $\pm$ 0.035 &{0.688 $\pm$ 0.039}&{0.761 {$\pm$ 0.019}}&0.720 $\pm$ 0.061\\
    celecoxib\_rediscovery & 0.713 $\pm$ 0.067 &0.567 $\pm$ 0.083&{0.802 {$\pm$ 0.029}}&0.860 $\pm$ 0.002 \\
    deco\_hop & 0.666 $\pm$ 0.044 &0.649 $\pm$ 0.025&0.733 {$\pm$ 0.109}&0.672 $\pm$ 0.118\\
    drd2 & 0.945 $\pm$ 0.007 &0.936 $\pm$ 0.016&{0.974 {$\pm$ 0.006}}&0.902 $\pm$ 0.117\\
    fexofenadine\_mpo & 0.784 $\pm$ 0.006 &{0.825 $\pm$ 0.019}&{0.856 {$\pm$ 0.039}}&0.806 $\pm$ 0.006\\
    gsk3b & 0.865 $\pm$ 0.043 &0.843 $\pm$ 0.039&0.881 {$\pm$ 0.042}&0.877 $\pm$ 0.055\\
    isomers\_c7h8n2o2 & 0.852 $\pm$ 0.036 &0.878 $\pm$ 0.026&{0.969 {$\pm$ 0.003}}&0.911 $\pm$ 0.031\\
    isomers\_c9h10n2o2pf2cl & 0.642 $\pm$ 0.054 &{0.865 $\pm$ 0.012}&{0.897 {$\pm$ 0.007}}&0.828 $\pm$ 0.126\\
    jnk3 & {0.783 $\pm$ 0.023} &0.702 $\pm$ 0.123&0.764 {$\pm$ 0.069}&0.785 $\pm$ 0.072\\
    median1 & 0.356 $\pm$ 0.009 &0.257 $\pm$ 0.009&{0.379 {$\pm$ 0.010}}&0.415 $\pm$ 0.001\\
    median2 & 0.276 $\pm$ 0.008 &{0.301 $\pm$ 0.021}&0.294 {$\pm$ 0.007}&0.408 $\pm$ 0.003\\
    mestranol\_similarity & 0.618 $\pm$ 0.048 &0.591 $\pm$ 0.053&0.708 {$\pm$ 0.057}&0.930 $\pm$ 0.106\\
    osimertinib\_mpo & 0.837 $\pm$ 0.009 &{0.844 $\pm$ 0.015}&{0.860 {$\pm$ 0.008}}&0.833 $\pm$ 0.011\\
    perindopril\_mpo & 0.537 $\pm$ 0.016 &0.547 $\pm$ 0.022&0.595 {$\pm$ 0.014}&0.651 $\pm$ 0.030\\
    qed & {0.941 $\pm$ 0.000} &{0.941 $\pm$ 0.001}&{0.942 {$\pm$ 0.000}}&0.947 $\pm$ 0.000\\
    ranolazine\_mpo & 0.760 $\pm$ 0.009 &{0.804 $\pm$ 0.011}&0.819 {$\pm$ 0.018}&0.810 $\pm$ 0.011\\
    scaffold\_hop & {0.560 $\pm$ 0.019} &0.527 $\pm$ 0.025&{0.615 {$\pm$ 0.100}}&0.529 $\pm$ 0.020\\
    sitagliptin\_mpo & 0.021 $\pm$ 0.003 &{0.582 $\pm$ 0.040}&0.634 {$\pm$ 0.039}&0.474 $\pm$ 0.085\\
    thiothixene\_rediscovery & 0.534 $\pm$ 0.013 &0.519 $\pm$ 0.041&{0.583 {$\pm$ 0.034}} &0.727 $\pm$ 0.089\\
    troglitazone\_rediscovery & {0.441 $\pm$ 0.032} &0.427 $\pm$ 0.031&{0.511 {$\pm$ 0.054}}&0.756 $\pm$ 0.141\\
    valsartan\_smarts & {0.178 $\pm$ 0.358} &0.000 $\pm$ 0.000&0.135 {$\pm$ 0.271}&0.000 $\pm$ 0.000\\
    zaleplon\_mpo & 0.358 $\pm$ 0.062 & {0.519 $\pm$ 0.029} &{0.552 {$\pm$ 0.033}}&0.499 $\pm$ 0.025\\
    \midrule
    Sum & 14.196 & {14.708} & {16.213} & {16.303} \\
    \bottomrule
  \end{tabular}
\end{table}

\clearpage
\section{SMILES from Figure~\ref{fig:similar-molecules-binaryfp}}
\label{apdx:smiles from figure}

Top pair:

\texttt{CCCCC}

\texttt{CCCCCCCCCCCCCCCCCCCC}

Bottom pair:

\texttt{CC1=CC=C(C=C1)C1=CC(=NN1C1=CC=C(C=C1)S(N)(=O)=O)C(F)(F)F}

\texttt{Cc1ccc(-c2cc(C(F)(F)F)nn2-c2ccc(-n3nc(C(F)(F)F)cc3-c3ccc(-n4nc(C(F)(F)F)cc4-c4ccc\\(S(N)(=O)=O)cc4)cc3)cc2)cc1}

\end{document}

%% file: math-notation.tex
\DeclareMathOperator*{\argmax}{arg\,max}

\newcommand{\X}{\ensuremath{\mathcal{X}}}
\newcommand{\R}{\ensuremath{\mathbb{R}}}
\newcommand{\D}{\ensuremath{\mathcal{D}}}
\newcommand{\prob}{\ensuremath{\mathbb{P}}}
\newcommand{\E}{\ensuremath{\mathbb{E}}}
\newcommand{\V}{\ensuremath{\mathbb{V}}}